# LLM-Agent-based Social Simulation for Attitude Diffusion

Deepak John Reji

*Department of Sociology, University of Limerick, Limerick, Ireland*

reji.deepak@ul.ie

**Keywords:** *agent-based modelling; social network dynamics; opinion polarisation; large language models; attitude formation; computational social science*

**Abstract**

This paper introduces [discourse_simulator](), an open-source framework that combines LLMs with agent-based modelling. It offers a new way to simulate how public attitudes toward immigration change over time in response to salient events like protests, controversies, or policy debates. Large language models (LLMs) are used to generate social media posts, interpret opinions, and model how ideas spread through social networks. Unlike traditional agent-based models that rely on fixed, rule-based opinion updates and cannot generate natural language or consider current events, this approach integrates multidimensional sociological belief structures and real-world event timelines. This framework is wrapped into an open-source Python package that integrates generative agents into a small-world network topology and a live news retrieval system. **discourse_sim** is purpose-built as a **social science research instrument** specifically for studying attitude dynamics, polarisation, and belief evolution following real-world critical events. Unlike other LLM Agent Swarm frameworks, which treat the simulations as a prediction black box, discourse_sim treats it as a theory-testing instrument, which is fundamentally a different epistemological stance for studying social science problems. The paper further demonstrates the framework by modelling the Dublin anti-immigration march on April 26, 2025, with N=100 agents over a 15-day simulation.

## 1. Introduction

We live in a world that is evolving rapidly technologically, socially, and politically. We have witnessed the evolution of AI, particularly the rise of autonomous systems, and the evolution of human society, and, at the same time, the unfolding of difficult realities: wars, humanitarian crises, natural disasters, and struggles for survival. In Ireland, particularly after 2023, we have seen an increase in racial incidents and, more recently, the largest immigration-related protest in the country. As an immigrant myself, understanding how societies respond to immigrants has always felt both important and personal. From the moment I arrived in Ireland, I have been motivated to understand how people act, react, and form perceptions about immigrants. With my background in data science, I felt a responsibility to apply what I have learned and acquired over the years to study these attitudes more deeply, and that motivation is at the heart of this research.

Public discourse following politically charged events is harder to explain through simple pairwise numerical exchanges. It often involves contextual depth, which involves heterogeneous ideological identities, multidimensional belief structures, exposure to real-time media, and peer influence mediated by social network position. Classical ABM formalisms, bounded confidence models, voter models, and cultural dissemination abstract all of this into a single floating-point opinion variable updated by deterministic rules. This abstraction trades realism for tractability.

The recent emergence of high-performance open-source LLMs creates an opportunity to populate ABMs with agents that reason in natural language, actively retrieve information, and generate posts that can be semantically scored. This new class of models, increasingly called Generative ABMs (GABMs) or generative social simulation, leverages LLMs to simulate human-like agents who can plan, reason, and interact via language (Liu et al. 2025). However, validation remains the central challenge; the flexibility that makes generative ABMs powerful also makes them difficult to validate, reproduce, and compare (Liu et al. 2025) also, there exists the fundamental question of the reliability of GABMs, whether a social phenomenon could be studied using these simulated hyper-realistic outputs.

No existing framework in Computational Social Science domain combines: (a) multidimensional belief decomposition across economic, cultural, security and humanitarian dimensions; (b) psychologically grounded inertia, conformity and emotional reactivity parameters; (c) empirically calibrated agent distributions; (d) live news retrieval via an Observe-Think-Act loop; and (e) a verified real-world event timeline, all in a reproducible, open-source package applicable to any critical event.

Hence, this research contributes to the following:

1. *Framework contribution*: discourse_simulator as a configurable, open-source generative ABM package
2. *Methodological contribution*: the Observe-Think-Act loop architecture that grounds LLM outputs in real retrieved news without requiring native tool-calling support
3. *Empirical contribution*: the first generative simulation of post-event immigration discourse dynamics in Ireland, with a verified 15-day event timeline

## 2. Background and Related Work

### 2.1 Generative Social Science and the ABM Tradition

Epstein argued that agent-based computational models enable a distinctively "generative" approach to social science to explain a social phenomenon, one must be able to grow it from the bottom up through decentralised local interactions of heterogeneous autonomous agents. His maxim, "if you didn't grow it, you didn't explain its emergence", remains the methodological touchstone for all ABM. (Epstein 1999). While earlier ABMs were criticised as "toy models" investigating abstract social dynamics (Larooij and Törnberg 2025), the field has matured significantly, with ABM now an accepted method across public health, criminology, migration, and sociology. (Macy and Willer 2002) (Gilbert and Terna 2000). The two key challenges which limited the adoption of ABMs in social sciences were the oversimplification of Human Behaviour (Lack of Realism) and the lack of Empirical Grounding (Validation and Calibration). Traditional models were struggling to replicate human reasoning, storytelling, learning, and cognitive biases. The Cognitive depth was missing, and the older models rarely accounted for the role of emotions, adherence to social norms, or the influence of memories and past interactions (Larooij and Törnberg 2025).

### 2.2 Opinion Dynamics Models: From Bounded Confidence to Social Influence

The formalisation of the belief update equations used in this framework is through bounded confidence models proposed independently by Hegselmann & Krause and by Deffuant et al. in 2000, which represent agents holding continuous opinions that they adjust through interaction, but only when interacting with agents sufficiently close in opinion. Deffuant, Neau, Amblard & Weisbuch (2001) Hegselmann & Krause (2002). The multidimensional belief model used in this framework directly addresses a limitation identified in the literature, that there is an urgent need for more theoretical work comparing and integrating alternative models, and the field suffers from a strong imbalance between theoretical studies and empirical work. Flache, Mäs, Feliciani et al. (2017) Castellano, Fortunato & Loreto (2009) Axelrod (1997)

### 2.3 LLM-Augmented Agent-Based Social Simulation

Park et al. (2023) fused large language models with computational interactive agents, introducing architectural and interaction patterns enabling generative agents to exhibit complex social behaviours like organising events, forming new acquaintances, and coordinating across time. The ability of these agents to simulate a believable human being with social roles opens up different possibilities to analyse different social situations. The opinion dynamics application in this framework could be most directly compared to Chuang et al. (2023)'s work, where they proposed LLM populations as an alternative to ABMs for opinion dynamics simulation and found a strong inherent bias in LLM agents towards producing accurate information, leading simulated agents toward consensus regardless of the personas they role-play. This is a critical limitation that the confirmation bias via psychological priors directly addresses. (Junchi yao et al.). Törnberg et al. (2023) used LLMs with ABM to explore the impact of news feed algorithms in simulated social media environments, which is similar to the live DuckDuckGo retrieval loop in this framework, which extends by grounding agents in real-time rather than synthetic news.

However, the introduction of LLMs exacerbates the long-standing validation tension in ABM. Their greater complexity heightens the difficulty of balancing standardisation with the flexibility that makes generative ABMs attractive. The verified event timeline and labelled evidence classification address this issue to an extent in this framework.

### 2.3 Computational Approaches to Immigration and Political Discourse

Entman's (1993) framing theory holds that news frames activate different cognitive schemas by selecting and emphasising certain aspects of a perceived reality. The four belief dimensions (economic, cultural, security, humanitarian) in this framework are direct computational operationalisations of the four dominant immigration frames identified in political communication research. Boomgaarden & Vliegenthart (2009). One of the important gaps this paper addresses is that most computational immigration discourse studies use classical NLP on real corpora (Twitter/X, news) but not generative simulation. The combination of generative agents + verified real-world event + live news retrieval is novel.

## 3. Framework Design

The package is a five-layer sequential pipeline. The user interacts exclusively with the discourse-simulator library, which orchestrates all internal components via SimConfig, which instantiates agents, the social network, LLM tools, and the news timeline in parallel. The engine executes the Observe-Think-Act (OTA) loop across T days × N agents. Belief updates revise the agent state after each posting round. Results are exposed as raw pandas dataframe.

### 3.1 Overview

At a chosen timestep (t == event_t), a priming event (e.g., "large anti-immigration protest in Dublin") is introduced. This shifts the tone of discourse and increases the salience of immigration in public debate. LLM-generated social media posts incorporate this event into agent communications. The simulation runs for 15 consecutive days (for eg. Day 0 = April 26, 2025; Day 14 = May 10, 2025 for the critical event), with each day representing one timestep in which every agent observes current news, reasons about it, produces a social media post, and updates their belief state based on post content and peer influence. The simulation produces both macro-level aggregate (average attitude, polarization, mood, exposure) and micro-level panel data (per-agent beliefs, messages, and scores across all 15 days).

| Parameter | Value | Description |
| --- | --- | --- |
| **Agents (N)** | 100 | Total synthetic social media users |
| **Timesteps (T)** | 15 | One day per timestep |
| **LLM Model** | mistral:7b-instruct-q4_0 | Local Ollama model for post generation |
| **Temperature (generation)** | 0.75 | Controls post diversity |
| **Temperature (scoring)** | 0.00 | Deterministic stance scoring |
| **Network topology** | Watts-Strogatz small-world | k=6 neighbours, p=0.3 rewiring |
| **Random seed** | 42 | Reproducibility of network and agent initialisation |

*Table 1: Simulation Parameters*

**Critical event & post-event timeline**

All agents share a single contextual backdrop: the critical events description. It is then injected into every agent prompt on every day of the simulation. This ensures posts are always grounded in the same event, while day-to-day variation is introduced by a verified 15-day timeline of real and contextually supported developments.

| Critical Event |
|---|
| In late April 2025, Dublin saw a large anti-immigration march from the Garden of Remembrance to the Custom House, with thousands protesting what they described as high levels of immigration and pressure on housing and public services. The protest featured slogans such as 'Ireland is Full' and 'Get Them Out', and it was met by a counter-protest ("United Against Racism"). The government has also faced pressure after a series of violent protests, arson threats, and resistance to building new asylum seeker centres in certain regions. This event has gained substantial media coverage and public attention. |

*Table 2: critical events and their description*

Each day in the post-event timeline is classified by one of three evidence levels to ensure methodological transparency. This classification prevents fabricated events from being treated as real stimuli and allows downstream analysis to distinguish agent reactions to verified versus inferred daily context.

### 3.2 Design Philosophy

Three principles that distinguish discourse_simulator from prior work:

1. Theoretical grounding over simplicity: ABMs carry multidimensional psychological profiles derived from established social psychology literature (Ajzen's theory of planned behaviour; dual-process models of attitude change). Every parameter has a theoretical referent.

2. Empirical calibration over an arbitrary distribution. Agent-type proportions are drawn from survey research (ESRI, LSE, Eurobarometer), not assumed equal. This positions the framework as empirically grounded.

3. Live grounding over static scenarios. The Observe-Think-Act loop retrieves real news on every simulation day, avoiding the criticism that generative ABMs operate in an informational vacuum.

### 3.3 Agent Architecture

Each agent is a stateful Python dataclass (Agent) with three categories of attributes: identity, multidimensional beliefs, and dynamic state. The agent population is initialised once before the simulation loop begins, with attribute values sampled from kind-specific prior distributions. The four agent kinds and their population proportions are grounded in Irish survey data and electoral research, not arbitrarily assigned:

| Kind | Proportion | Empirical Basis | Attitude Prior |
| --- | --- | --- | --- |
| centrist | 45% | Dominant group in Irish politics; housing-anxious but not ideologically far-right (Eurobarometer 2025) | Uniform(-0.4, +0.4) |
| pro_imm | 25% | Younger cohort, NGO workers, immigrant community; under-35s significantly more pro-immigration per ESRI surveys | Uniform(-1.0, -0.3) |
| far_right | 20% | ~52–59% of Irish adults say immigration 'too high' (LSE/ESRI 2024), but hard far-right identity is far narrower than anti-immigration sympathy | Uniform(+0.5, +1.0) |
| media | 10% | Journalists, institutional commentators; modelled as lower reactivity and higher openness | Uniform(-0.3, +0.3) |

*Table: Agent types and distribution*

The distribution is centrist-heavy and asymmetric. An even four-way split would misrepresent the actual Irish public opinion landscape in 2025, where far-right identity is a minority position despite broad scepticism about immigration levels.

| Attribute | Type | Values | Description |
| --- | --- | --- | --- |
| id | str | agent_0 … agent_99 | Unique agent identifier |
| kind | str | far_right \| pro_imm \| centrist \| media | Ideological type: determines prior distributions |
| attitude | float | [-1.0, +1.0] | Core immigration stance. -1 = strongly pro-immigration; +1 = strongly anti-immigration. Updated every timestep. |
| exposure | float | [0.0, 1.0] | Cumulative exposure to threat-framed immigration narratives. Monotonically increases when threat-coded news is encountered. |
| quirk | str | 8 options | Assigned once at first post. Determines writing style (e.g., sarcasm, emojis, hashtags, formal tone). |

*Table: identity attributes*

Attitude is driven by four sub-belief dimensions, each independently updated and combined into a composite belief score. This separates the channels through which different types of news affect the overall stance.

| Attribute | Init Range by Kind | Description |
| --- | --- | --- |
| economic_threat_belief | far_right: [0.4, 0.9] pro_imm: [-0.5, 0.1] centrist: [-0.2, 0.4] | Degree to which agent believes immigration causes economic harm (job competition, housing pressure) |
| cultural_threat_belief | far_right: [0.4, 0.9] pro_imm: [-0.5, 0.0] centrist: [-0.2, 0.3] | Degree to which agent perceives cultural threat from immigration |
| security_threat_belief | All: derived from exposure | Belief that immigration is linked to crime or security risks. Updated by threat-coded news salience. |

| Attribute | Range | Role in Simulation |
|---|---|---|
| humanitarian_belief | far_right: [-0.5, 0.1] pro_imm: [0.4, 1.0] centrist: [0.0, 0.5] | Degree to which agent values humanitarian obligation to migrants. Updated by humanitarian-framed news. |

*Table: Belief dimension*

The openness × inertia coupling directly operationalises psychological anchoring to simulate seemingly realistic behavior (Petty & Cacioppo (1986), Elaboration Likelihood Model; Festinger (1957) cognitive dissonance) and the emotional_reactivity parameter operationalises affective intelligence theory (Marcus, Neuman & MacKuen (2000) Affective Intelligence and Political Judgment)

| Attribute | Range | Role in Simulation |
|---|---|---|
| openness | [0.1, 1.0] (kind-stratified) | Controls belief inertia. High openness = less resistance to attitude change. Feeds directly into the inertia multiplier: inertia = 1 - openness * 0.5 |
| conformity | [0.3, 0.8] | Scales how strongly the agent's attitude is pulled toward peer_mean. High conformity agents cluster around their social network. |
| emotional_reactivity | [0.2, 1.0] (kind-stratified) | Multiplies the impact of threat-coded and humanitarian news on sub-belief dimensions. |
| trust_peers | [0.4, 0.9] | Scales the peer influence pull. Multiplied by conformity to compute net social pressure. |

*Table: Psychological Profile Attributes*

| Attribute | Initial Value | Update Rule |
|---|---|---|
| mood | 0.0 | mood(t) = 0.8 * mood(t-1) + shock, where shock = -0.1 if threat news present, +0.04 otherwise. Decays toward 0 over time. |
| messages | [] | Appended each timestep with the agent's generated post. Last 5 posts used as memory context. |
| attitude_history | [] | Appended each timestep with the updated attitude value. Provides full trajectory for analysis. |
| reasoning_log | [] | Appended each timestep with tool call trace (tool name + input). Enables audit of agentic reasoning. |

### 3.4 Agentic Observe-Think-Act Loop

Post generation is agentic: on each day, every agent executes a structured three-phase reasoning loop before producing their social media post. This avoids static template-based generation and grounds each post in live-retrieved news and the agent's own evolving memory.

The original implementation was built using LangGraph's create_react_agent framework for tool orchestration. However, mistral:7b-instruct-q4_0 does not natively support the OpenAI-compatible tool-calling schema required by these frameworks (.bind_tools()). All agents immediately hit exceptions. The framework was replaced with a manual Python-orchestrated loop that preserves full agentic behaviour without delegating tool-call parsing to the model.

This has been more architecturally feasible for this framework with locally-run LLMs compared to pure ReAct frameworks because ReAct (Yao et al. 2022) requires the LLM to parse tool-call schemas and this fails for small instruction-tuned models like Mistral 7B. Besides the manual OTA loop decouples tool execution (Python) from reasoning (LLM), making the framework model-agnostic and reproducible.

**Phase 1: Observe (Tool Calls)**

Python directly invokes the tools and collects their outputs as observations:

| Tool | Input | Output | Purpose |
| --- | --- | --- | --- |
| search_immigration_news | Query string derived from agent kind + day context | Up to 1,200 characters of DuckDuckGo search results anchored to 'Dublin immigration march April 2025 Ireland {query}' | Provides real-world news context beyond the static POST_EVENT_TIMELINE |
| recall_agent_memory | JSON array of agent's last 5 posts | Natural language summary of posting history and most recent post (first 100 chars) | Ensures day-to-day narrative consistency; prevents attitude flips |
| get_sentiment_of_text | Any text string | pro_immigration / anti_immigration / neutral with lexicon-based confidence score | Available to agents as an optional analytical tool; also used in standalone scoring |

**Phase 2: Think + Act (LLM Call)**

All observations from Phase 1, the agent's full profile, verified news entry, and the permanent critical event context are concatenated into a single structured prompt and passed to the Ollama LLM. The prompt specifies the agent's identity attributes, psychological profile, belief dimensions, mood, and writing quirk, and explicitly instructs the model to evolve stance gradually without overnight flips. The model outputs a single social media post of a maximum of 40 words

> **post_t = LLM(CRITICAL_EVENT + profile + memory + search_result + todays news)**

**Attitude Scoring (Post Interpretation)**

After all agents post on a given day, each post is converted to a numeric attitude score in [-1, +1] using a secondary LLM call with temperature = 0.0 (deterministic). This score represents how anti-immigration the post is, independently of the agent's internal attitude state.

**LLM-Based Scoring**

The scoring prompt presents each post to the model and requests a single float on the scale:

| Score | Interpretation |
| --- | --- |
| -1.0 | Strongly pro-immigration (e.g., solidarity messaging, refugee rights advocacy) |
| 0.0 | Neutral or ambivalent (e.g., reporting facts without stance) |
| +1.0 | Strongly anti-immigration (e.g., 'Ireland is Full', deportation demands) |

The model output is parsed with a regex for the first float-like token (pattern: -?\d+\.?\d*). The result is clipped to [-1, +1] using numpy.clip. If parsing fails, the score defaults to 0.0.

### 3.5 Computations

**Belief Update Mechanism**

At the end of each timestep, after all posts have been generated and scored, each agent's belief state is updated through a five-component model. The update integrates news-driven belief salience, peer network influence, psychological inertia, and mood dynamics.

**News Salience Computation**

Each day's news entry is scanned for threat-coded and humanitarian-coded keywords. The counts are multiplied by a fixed salience coefficient (0.06 per keyword occurrence) and then by the agent's emotional_reactivity multiplier:

| Signal | Keywords | Coefficient |
|---|---|---|
| news_threat | arson, attack, violence, crime, danger, get them out, deportation, deport | count × 0.06 × emotional_reactivity |
| news_humanitarian | refugee, asylum, rights, children, family, compassion, solidarity, waiting | count × 0.06 × emotional_reactivity × openness |

Threat salience increases security_threat_belief. Humanitarian salience increases humanitarian_belief, but is additionally modulated by openness; agents with low openness are less receptive to humanitarian framing.

**Peer Influence**

Each agent's social neighbourhood is defined by the Watts-Strogatz graph. The mean attitude score of all neighbours on the current day is computed, and a peer pull force is applied proportional to the agent's conformity and trust_peers:

> **peer_pull = conformity × trust_peers × (peer_mean - attitude_t)**

If an agent has no neighbours with scores in the current timestep, peer_pull = 0. The force is directional: agents above the neighbourhood mean are pulled downward; agents below are pulled upward. The magnitude is bounded by conformity × trust_peers (maximum ≈ 0.64 for maximum values of both). **(**McPherson, Smith-Lovin & Cook 2001)

**Mood Dynamics**

Mood is a short-term affective state that decays toward zero and is shocked by news framing:

$$\text{mood}(t) = \text{clip}(\ 0.8 \times \text{mood}(t-1) + \text{shock},\ -1,\ +1\ )$$

The shock term is -0.1 when threat-coded news is present (news_threat > 0), and +0.04 otherwise. The decay coefficient 0.8 means a mood shock dissipates to below 10% of its original magnitude within approximately 10 days, consistent with psychological literature on affective adaptation.

**Belief Inertia**

Psychological anchoring is modelled as an inertia multiplier derived from the agent's openness. High-openness agents change more readily; low-openness agents resist change:

$$\text{inertia} = 1 - \text{openness} \times 0.5 \quad (\text{range: } [0.5, 1.0])$$

An agent with openness = 0.0 (minimum) has inertia = 1.0, meaning their attitude is entirely determined by their prior state. An agent with openness = 1.0 (maximum) has inertia = 0.5, meaning prior state and new information contribute equally. (Eagly & Chaiken 1993)

**Composite Belief Score**

The four belief dimensions are combined into a single composite score that exerts pressure on attitude:

$$\text{belief\_composite} = 0.3 \times \text{economic\_threat} + 0.3 \times \text{cultural\_threat} + 0.2 \times \text{security\_threat} - 0.2 \times \text{humanitarian}$$

Economic and cultural threat beliefs are weighted equally at 0.3 (together accounting for 60% of the composite), reflecting their dominant role in Irish immigration discourse in 2025. Security threat and humanitarian beliefs each contribute 20%, with humanitarian beliefs exerting a negative (pro-immigration) pressure.

### 3.6 Full Attitude Update

The final attitude update integrates all five components:

$$\text{attitude}(t) = \text{clip}(\text{inertia} \times \text{attitude}(t-1) + (1 - \text{inertia}) \times (0.4 \times \text{own\_score} + 0.3 \times \text{peer\_pull} + 0.3 \times \text{belief\_composite}),\ -1,\ +1\ )$$

Within the flexible component (weighted by 1 - inertia), own_score (the LLM-interpreted score of the agent's own post today) has the highest weight at 0.4, reflecting self-reinforcement through public expression. Peer pull and belief composite each contribute 0.3. All outputs are clipped to [-1, +1].

**Exposure Update**

Exposure accumulates monotonically when threat-coded news is present:

$$\text{exposure}(t) = \min(1.0, \text{exposure}(t-1) + 0.07 \times \text{emotional\_reactivity}) \quad \text{if news\_threat} > 0$$

The coefficient 0.07 ensures that even the most emotionally reactive agent (emotional_reactivity = 1.0) requires approximately 14 threat-coded days to saturate from zero exposure to 1.0, which is consistent with the simulation window length.

*Entman (1993) Framing: Toward clarification of a fractured paradigm, Journal of Communication 43(4); Boomgaarden & Vliegenthart (2009) How news content influences anti-immigration attitudes, European Journal of Political Research*

*ABM Foundations*

- *Epstein, J.M. (1999). Agent-based computational models and generative social science. Complexity, 4(5), 41–60.*

- *Epstein, J.M. & Axtell, R. (1996). Growing Artificial Societies. Brookings.*

- *Macy, M.W. & Willer, R. (2002). From factors to actors: Computational sociology and agent-based modeling. Annual Review of Sociology, 28, 143–166.*

- *Gilbert, N. & Terna, P. (2000). How to build and use agent-based models in social science. Mind & Society, 1, 57–72.*

- *Squazzoni, F. (2010). The impact of agent-based models in the social sciences after 15 years. History of Economic Ideas, 18(2).*

*Opinion Dynamics*

- *Deffuant, G., Neau, D., Amblard, F. & Weisbuch, G. (2001). Mixing beliefs among interacting agents. Advances in Complex Systems, 3(1–4), 87–98.*

- *Hegselmann, R. & Krause, U. (2002). Opinion dynamics and bounded confidence models. JASSS, 5(3).*

- *Axelrod, R. (1997). The dissemination of culture. Journal of Conflict Resolution, 41(2), 203–226.*

- *Flache, A., Mäs, M., Feliciani, T. et al. (2017). Models of social influence: Towards the next frontiers. JASSS, 20(4).*

- *Castellano, C., Fortunato, S. & Loreto, V. (2009). Statistical physics of social dynamics. Reviews of Modern Physics, 81(2), 591–646.*

*LLM-ABM*

- *Park, J.S., O'Brien, J.C., Cai, C.J. et al. (2023). Generative Agents: Interactive Simulacra of Human Behavior. UIST 2023.*

- *Chuang, Y.-S., Goyal, A., Harlalka, N. et al. (2023). Simulating Opinion Dynamics with Networks of LLM-based Agents. NAACL-HLT 2024.*

- *Törnberg, P., Valeeva, D., Uitermark, J. et al. (2023). Simulating Social Media Using Large Language Models to Evaluate Alternative News Feed Algorithms. arXiv:2310.05984.*